\UseRawInputEncoding
\documentclass[11pt]{amsart}
\usepackage{geometry}                
\usepackage{graphicx}
\usepackage{amssymb}
\usepackage{epstopdf}
\title{A fuzzy take on the logical issues of statistical hypothesis testing}
\author{Matthew Booth and Fabien Paillusson}

\date{%
\today \\ Affiliation: School of Mathematics and Physics, University of Lincoln, UK}                                           

\begin{document}
\begin{abstract}Statistical Hypothesis Testing (SHT) is a class of inference methods whereby one makes use of empirical data to test a hypothesis and often emit a judgment about whether to reject it or not. In this paper we focus on the logical aspect of this strategy, which is largely independent of the adopted school of thought, at least within the various frequentist approaches. We identify SHT as taking the form of an unsound argument from Modus Tollens in classical logic, and, in order to rescue SHT from this difficulty, we propose that it can instead be grounded in t-norm based fuzzy logics. We reformulate the frequentists' SHT logic by making use of a fuzzy extension of modus Tollens to develop a model of truth valuation for its premises. Importantly, we show that it is possible to preserve the soundness of Modus Tollens by exploring the various conventions involved with constructing fuzzy negations and fuzzy implications (namely, the S and R conventions). We find that under the S convention, it is possible to conduct the Modus Tollens inference argument using Zadeh's compositional extension and any possible t-norm. Under the R convention we find that this is not necessarily the case, but that by mixing R-implication with S-negation we can salvage the product t-norm, for example. In conclusion, we have shown that fuzzy logic is a legitimate framework to discuss and address the difficulties plaguing frequentist interpretations of SHT. \end{abstract}
\maketitle
\section{Introduction}
\label{sec1}

According to Popper  \cite{Popper}, a theory can be deemed scientific if it implies the {\it impossibility} of certain events that are otherwise perfectly conceivable (or even expected). A test of a scientific theory should thereby be an attempt at falsification by means of observing such impossible events. To many observers this refutability strategy was an important step towards solving the induction problem that plagues the philosophy of science and, at the same, undermines the legitimacy of scientific claims.  While the Popperian view on scientific practice is far from being the final say on what constitutes a scientific theory \cite{Chalmers1999}, it is nevertheless the most popular and invoked philosophy of science in scientific circles. From a logical standpoint the rejection argument for a hypothesis $H$ given evidence $E$ operates via the {\it Modus Tollens} (MT) inference rule: 
\begin{eqnarray}
&  H \Longrightarrow \lnot E  \\
&  E  \\
   \cline{2-3}
\therefore &  \lnot H 
\end{eqnarray}where $\lnot$ represents the negation symbol and $\Longrightarrow$ an entailment. According to this logic, a solitary observation of a falsifying event $E$ necessarily undermines the hypothesis $H$. In practice, however, since there is no such thing as a 'pure observation', a single counter-instance does not falsify the theory. There are at least two reasons why this is the case. First, the research hypothesis $H$ being tested is necessarily supplemented by a set of auxilliary hypotheses $H'$ (i.e. underlying 'hidden' assumptions) that connect it to the real world. What is actually being tested, therefore, is the conjunction $H\wedge H'$. This is an example of holistic underdetermination; one can falsify this conjunction but not the research hypothesis alone \cite{Duhem}.  Second, experimental data is always susceptible to statistical uncertainties and it is unclear in what sense MT can still hold in such cases. This question will be the central theme of the present paper. That evidence is subject to statistical effects was already known and appreciated in the last decades of the 19$^{th}$ century as an integral part of scientific practice. However, this was not addressed quantitatively until the the beginning of the 20$th$ century with Student's exact derivation of the $t$-distribution in probability theory, and was turned into a more general framework by Fisher, Neynman and Pearson in the 1920s \cite{Lehmann1993}. In what follows we shall refer to this alleged general framework as {\it Statistical Hypothesis Testing} (SHT). To prevent any misunderstanding, we wish to clarify that we use SHT here as an umbrella term for the various classical, frequentist methodologies and philosophies pertaining to hypothesis testing under statistical considerations. As a result, the present study does not address the developments in hypothesis testing that rely on Bayesian inference. The interested reader may explore some of the tools, challenges and promising avenues of Bayesian hypothesis testing in Refs. \cite{Tendeiro19, Stern17, Stern14, Lu20}. The main frequentist schools we shall consider under the SHT term are Fisher's \cite{Fisher55}, Neyman \& Pearson's \cite{Neyman1928} and the {\it Null Hypothesis Significance Testing} (NHST) school; the latter often being considered as the illegitimate child of the two former \cite{Perezgonzalez15}. Despite their very strong philosophical differences, it is our appreciation that these three schools of thought can be summarised by a reasoning of the form
\vspace{-2mm}
\begin{eqnarray}
     {\rm P1}: &  H \Longrightarrow (P(E|H) \geq \alpha)  \label{eq2}\\
     {\rm P2}:  & \lnot (P(E|H) \geq \alpha) \label{eq3} \\
    \cline{2-2}
     \therefore & \:\:\:\:\:\:\:\: \lnot H \label{eq4}
\end{eqnarray}
In the above argument, $P(E|H)$ is the {\it a priori} knowable conditional probability of some observed evidence $E$ given an assumed true hypothesis $H$. ${\rm P1}$ should therefore be interpreted as a proposition about $E$: ${\rm P1}\equiv$ "If $H$ then $E$ is such that $P(E|H) > \alpha$". Likewise ${\rm P2}$ is also about the observed evidence $E$: ${\rm P2}\equiv$ "$E$ is such that $P(E|H) < \alpha$". The quantity $P(E|H)$ is the so-called $P$ value of the observed evidence and is a commonly used frequentist measure in medical studies and life sciences. Different schools attribute different meaning and value to the reasoning in Eqs.\eqref{eq2}-\eqref{eq4}. For example, 'Fisherians' would hold that a hypothesis can eventually be intellectually rejected under certain circumstances via an inductive reasoning while 'Neymanians' would contend that the hypothesis itself can usually not be rejected but a research worker may behave {\it as if} it was rejected \cite{Greco11}.  
Of the three schools we are discussing here, NHST is by far the most predominant methodology being used nowadays and as such has received many critics on two principal accounts: a) its methodological framework comes as an incomplete and simplified mix of Fisher's and Neyman \& Pearson's philosophies which does not take into account the more subtle but necessary attributes of either of these respective schools \cite{Perezgonzalez15}. b) As a result, NHST practitioners often misunderstand and mischaracterise the meaning of the numbers they obtain through the application of NHST's various statistical procedures \cite{Nickerson00, Gigerenzer04}. A commonly held false claim from a substantial portion of NHST practitioners is that their methods enable them to know the probability for the chosen hypothesis to be false \cite{Gigerenzer04}, {\it i.e.} from the reasoning in Eqs.\eqref{eq2}-\eqref{eq4} some NHST practitioners would claim that they know $P(H|E)$.
As pointed out by Goodman \cite{Goodman99}, this is of course fallacious. From Bayes' rule we have the following relationship between $P(H|E)$ and $P(E|H)$:
\begin{equation}
P(H|E) = \frac{P(E|H)P(H)}{P(E|H)P(H)+P(E| \lnot H)(1-P(H))} \label{eq1}
\end{equation}where both the prior $P(H)$ and the conditional probability $P(E| \lnot H)$ for $E$ to be observed given $H$ being false have unknown values. Goodman contends then that:
\begin{enumerate}
\item Evaluating how likely is $H$ given the observed evidence $E$ requires knowing $P(H|E)$,
\item Evaluating $P(H|E)$ requires the knowledge of two additional variables $P(H)$ and $P(E| \lnot H)$ which are independent from $P(E|H)$,
\item It is therefore not possible to reject $H$ on the sole basis of observing $E$ and knowing {\it a priori} $P(E|H)$.
\end{enumerate}
This conclusion can be illustrated by plotting the probability density map of $P(H|E)$ for $P(E|H)=0.04$ ({\it i.e.} usually considered as statistically significant in medical statistics) with $P(H)$ and $P(E| \lnot H)$ taking all possible values between $0$ and $1$.
\begin{figure}[h!]
\centering
\includegraphics[width=0.5\linewidth]{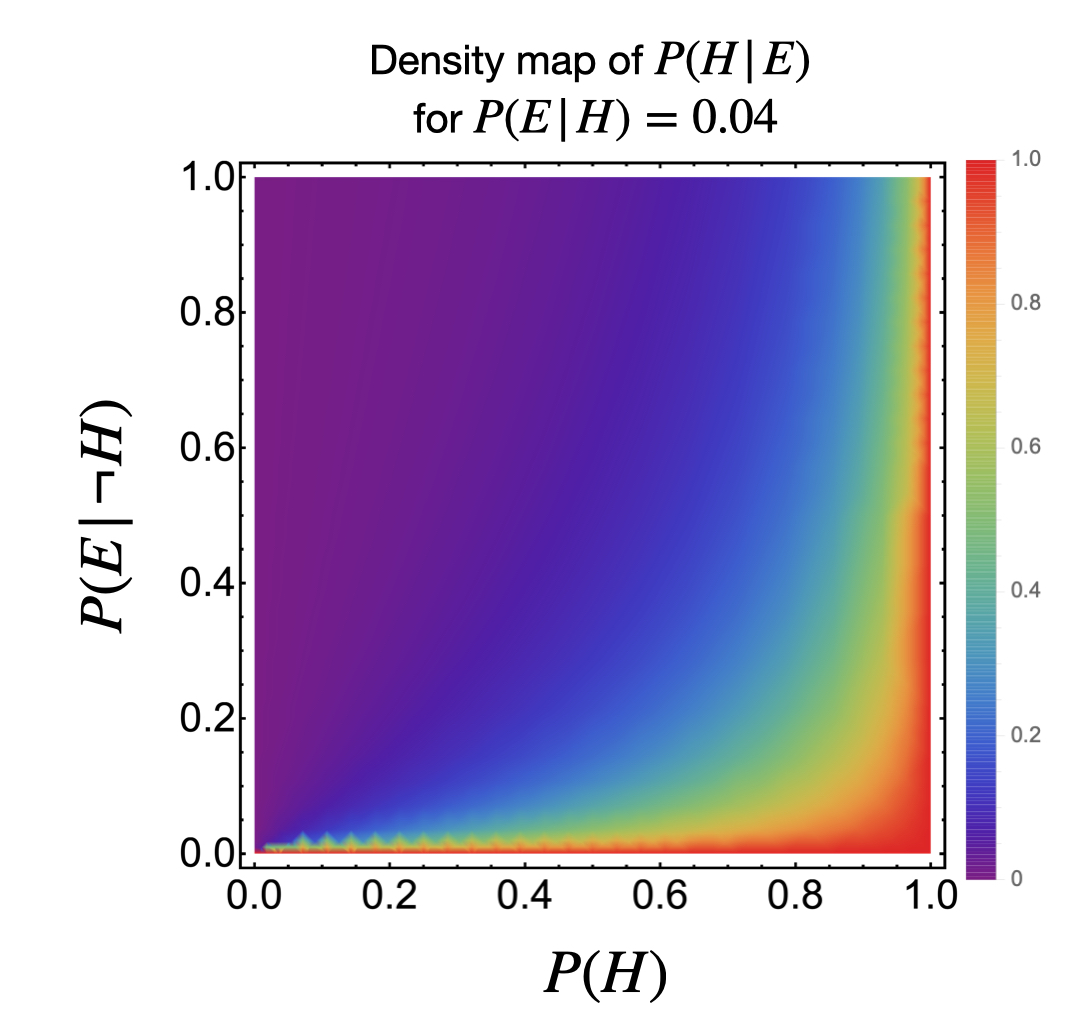}
\caption{Density map plot of $P(H|E)$ for $P(E|H) = 0.04$ for $P(H)$ and $P(E| \lnot H)$ taking independently all possible values between $0$ and $1$. Colours leaning towards purple refer to low probability values while redder colours correspond to probabilities close to 1.}
\label{fig1}
\end{figure}In Fig.\ref{fig1}, we see that despite the fact that $P(E|H) = 0.04$, about half of the parameter space (most of the bottom right corner) would give more than $20 \%$ chance $H$ to be true given $E$. In this light, it becomes inevitable to see hypothesis rejection on the sole basis of $P$ value as an invalid, fallacious argument.
This problem - that what could be inferred from a reasoning of the type \eqref{eq2}-\eqref{eq4} was not quantified by an actual probability - was already known and pondered by pioneers of SHT. Fisher \cite{Lehmann1993} would admit that:
\begin{quotation}
...more generally, however, a mathematical quantity of a different kind, which I have termed mathematical likelihood, appears to take its place [i.e. the place of probability] as a measure of rational belief when we are reasoning from the sample to the population.
\end{quotation}When the hypothesis $H$ can be specified as one or more parameter values in a model then the likelihood $\mathcal{L}(H|E) \equiv P(E|H)$ is the quantity that seems to be recommended for rational reasoning here. Neyman \& Pearson \cite{Neyman1928} shared a similar opinion
\begin{quotation}
Here and later the term "probability" used in connection with Hypothesis A [i.e. that a sample data is a finite realisation of a theoretical population] must be taken in a very wide sense. It cannot necessarily be described by a single numerical measure of inverse probability; as the hypothesis becomes "less probable" our confidence in it decreases, and the reason for it lies in the meaning of the particular contour system that has been chosen [i.e. critical regions].
\end{quotation}In the above citation, texts in square brackets were added to provide missing context. This weakened version of probability to make a decision about $H$ is contrary to Bayesian inference whereby the only thing newly obtained data can permit is to update a pre-existing belief about $H$ quantified by $P(H)$. Furthermore, this prior is not enough for the update to occur as indicated in Fig. \ref{fig1}. This constitutes a traditional critique of Bayesian approaches in that they are claimed to rely on the subjectivity of the practitioner. Such a blanket critique of Bayesian strategies on the sole basis that prior distributions are needed is, however, unwarranted on two grounds. First, it is illusory to think that non-Bayesian analyses escape subjectivity since they always require individual expert knowledge to set the hypotheses, alternative hypotheses and $\alpha$-levels of significance. Second, it dismisses a fruitful research programme in Bayesian analysis that seeks to develop objective priors by generalising and formalising notions such as the principle of indifference and the principle of insufficient reason \cite{Consonni18}. How these could be used in the case of Bayesian hypothesis testing is discussed for example in \cite{Stern14}. In principle, the statistician is thus faced with a conundrum. They can either use a frequentist approach and be criticised on the grounds that their reasoning can never give the --- assumed necessary --- probability for the hypothesis to be false given the evidence, or they may use a Bayesian approach and, other issues aside, be asked to justify its objectivity. The present paper is concerned only with the former case, since this is the more heavily used practice in contemporary research. Thus, even if one assumes that the likelihood, or some wider notion of probability, may be key to expressing a valid pseudo-inductive judgment, the question we wish to address in the present study remains: does the reasoning evoked in Eqs. \eqref{eq2}-\eqref{eq4} hold-up to that promise? If one works with a logic where MT holds, then this argument is valid by construction. Whether the argument is sound or not, {\it i.e.} whether both ${\rm P1}$ and ${\rm P2}$ are true or not, is more questionable. If we are forced to classify ${\rm P1}$ as being either true or false then it is must necessarily be false since, under $H$ being true, any $E$ with non-zero probability can in principle be observed, not only those with a probability higher than a set threshold. Thus, from a classical logic point of view, basing the rejection of a hypothesis solely on the $P$ value is also fallacious. That there cannot exist a {\it Probabilistic Modus Tollens} was also put forward by Sober \cite{Sober1999} and Greco \cite{Greco11} whom essentially echo a sentiment by Neyman \& Pearson \cite{Neyman1928}
\begin{quotation}
It is indeed obvious, upon a little consideration, that the mere fact that a particular sample may be expected to occur very rarely in sampling from $\Pi$ [i.e. the hypothetical population] would in itself justify the rejection of the hypothesis that it had be so drawn, if there were no more probable hypotheses conceivable.
\end{quotation}This line of argument in the philosophy of SHT \cite{Neyman1928, Sober1999, Greco11} considers that the MT reasoning developed in Eqs. \eqref{eq2}-\eqref{eq4} is basically fallacious but can be rectified by considering $H$ as being rejected in favour of an alternative, more likely hypothesis, rather than being rejected as a standalone hypothesis. While this strategy is clearly trying to infer the most out of the available data \cite{Neyman1933}, we believe that it does not permit SHT as a reasoning strategy to be grounded in classical logic. More specifically, one cannot claim that the reasoning in Eqs. \eqref{eq2}-\eqref{eq4} is fallacious and state at the same time that a judgement based on likelihood ratios is valid without contradicting themselves. The reason for this is that the likelihood ratio is still subject to statistical effects and one still needs to evaluate the corresponding probability for this ratio to lie in a certain critical region; which is equivalent to the reasoning in Eqs. \eqref{eq2}-\eqref{eq4}. As Neyman \& Pearson put it \cite{Neyman1928}
\begin{quotation}
It would be possible to use the [likelihood] ratio $\lambda$ as a criterion but this without the knowledge of $P_{\lambda}$ does not enable us to estimate the extent of the form (1) error.
\end{quotation}The form (1) error mentioned being the frequency rate with which one may wrongly reject the tested hypothesis upon repeating many times the same experiment. ideally, one would want this form (1) error to be as small as possible (usually set by the value $\alpha$ in eq. \eqref{eq2}) under other specific constraints set by the experimenter (e.g. minimizing other forms of judgement error). \newline Based on the above discussion we agree with Sober \cite{Sober1999} that "there is no probabilistic analog of modus tollens" but we have also shown that a) this fundamental state of affairs could not be salvaged by simply comparing hypotheses and b) the logical problem has been shown to occur when propositions need to be exactly true or false. We cannot deduce from this restricted analysis that there is no logic within which SHT can make sense. Indeed, in the present work, we explore the possibility of grounding SHT ---  as described in Eqs. \eqref{eq2}-\eqref{eq4} --- in infinitely many-valued logic rather than in conventional classical logic. Section \ref{sec2} is dedicated to present and motivate many-valued logics. Section \ref{sec3} discusses how certain formulations of fuzzy logic might enable SHT to be formulated as a valid and sound argument. Finally, we evaluate more closely which of the various conventions surrounding fuzzy negation and fuzzy implication provide a viable logic with which to ground $P$ value reasoning and discuss where do our findings leave us with regards to scientific reasoning in Section \ref{sec4}.

\section{Classical logic, multivalued logic and fuzzy logic}
 \label{sec2}
\begin{figure}[h!]
\centering
\includegraphics[width=0.5\linewidth]{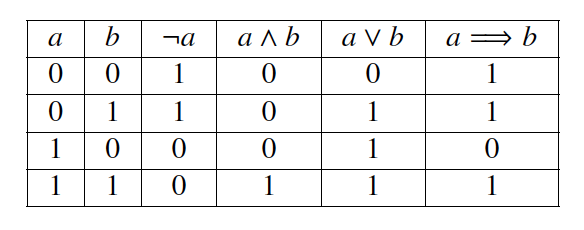}
\caption{Truth table of classical logic.}
\label{fig2}
\end{figure}

Classical logic or propositional logic is a formal language which comprises: a set of atomic propositions $D=(a,b,c,...)$; a set of connectives $C=(\neg,\vee,\wedge, \Longrightarrow)$, where $\neg$ stands for {\it negation}, $\vee$ for {\it disjunction}, $\wedge$ for {\it conjunction} and $\Longrightarrow$ for {\it implication}, that bind multiple atomic propositions into sentences; and a valuation function $\nu:C \mapsto \{0,1\}$, which assigns a truth value of either $1$ or $0$ to the connectives depending on the truth value of the propositions they bind (cf. table in Fig. \ref{fig2}). As we have seen in Section \ref{sec1}, it is this last feature that creates a problem for SHT; the MT inference rule in classical logic yields perfectly true and exact conclusions only to the extent that the premises themselves are perfectly true and exact. Yet, as we have discussed, the premises involved in the $P$ value reasoning from Eqs. \eqref{eq2}-\eqref{eq4} are necessarily false. For example, if $\alpha =0.05$ in premise ${\rm P1}$ and one interprets $E$ as $S>s$ (where $S$ is some aggregate random variable and $s$ its realisation), then one can conceive of realisations $s$ ({\it i.e.} models) where $H$ is true but $P(S>s|H) < 0.05$, thereby automatically making ${\rm P1}$ false. 

What if, however, one departs from the idea that propositions must necessarily be either true or false by adopting shades of truth? Allowing such a middle ground between true and false to exist might help model the fact that one may credit a given proposition as being, for example, "more true than false" (albeit not strictly true). 
The desire to apply logic to situations where propositions are perceived as partially true (or indeed where some things are "truer" than other things \cite{Smets87}) has motivated the development of many-valued logics (MVLs) --- logics which accommodate truth values that are intermediate between true and false. The concept of MVL is far from new and can be traced at least as far back as the 4-valued logics used in Buddhist and Greek philosophies under the names of {\it Catuskoti} and {\it Tetralemma}, respectively \cite{Priest10}. Of course, adopting a number-based representation of truth values enables a greater diversity of MVLs. The principle of bivalence that characterises classical logic is rejected within an MVL, and the valuation function $\nu$ may thus assign a truth value that is between $0$ and $1$. The truth value of a sentence is still contingent upon the truth values of the propositions that the connectives bind, but the degree of truth associated with the atomic propositions themselves are no longer two-valued. 

\begin{figure}[h!]
\centering
\includegraphics[width=1\linewidth]{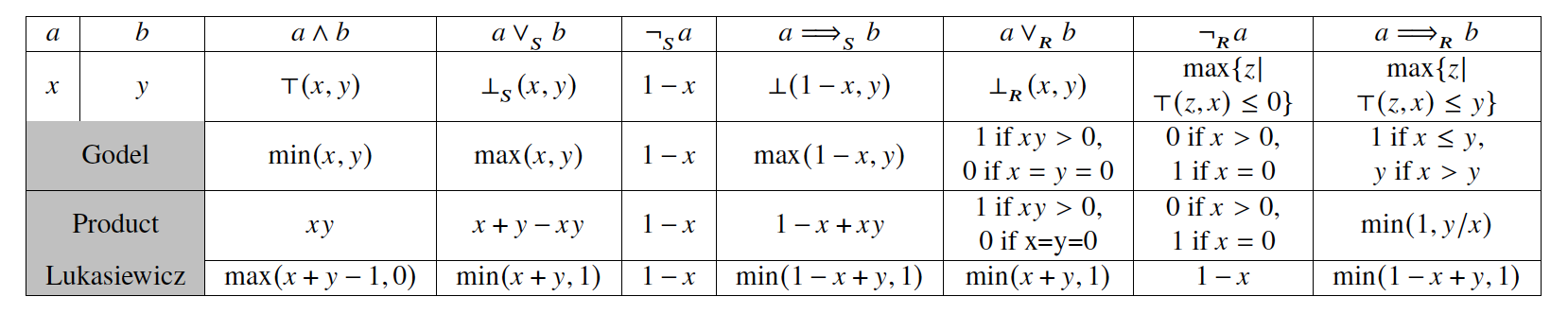}
\caption{Truth table of t-norm based IMVLs for the three most common t-norms.}
\label{fig3}
\end{figure}

Infinitely many-valued logics (IMVLs) are a subclass of MVLs in which the valuation function maps to a continuous variable on the interval $[0,1]$, such that there are infinitely many truth values that a proposition could be assigned. The most widely used IMVLs rely on so-called t-norms $\top(x,y)$ between two numbers $x,y \in [0,1]$. A t-norm is a binary operator which has the following properties: \begin{enumerate}
\item $\top(x,y)=\top(y,x);$
\item $\top(x,y) \leq \top(w,z)$ if $x\leq w$ and $y\leq z$;
\item $\top(x,\top(y,z)) =\top(\top(x,y), z)$; and
\item $\top(x,1) =x$\:.
\end{enumerate} In the table shown in Fig. \ref{fig3} the third column represents the conjunction operator whose valuation is identified to a t-norm in all t-norm based IMVLs. Columns 4-6 show how the conditional, negation and disjunction connectives are defined under the S-implications convention: a strong negation is defined as $\nu(\lnot_{_S} a) \equiv 1-\nu(a)$, the corresponding S-implication is defined from $a \Longrightarrow_{_S} b \equiv (\lnot_{_S} a \vee_{_S} b)$, and the S-disjunction is defined via the t-conorm $\nu(a \vee_{_S} b) \equiv \bot_{_S}(\nu(a),\nu(b)) \equiv 1- \top(\nu(\neg_{_S} a),\nu(\neg_{_S} b))$. Columns 7-9 introduce the negation, disjunction and conditional connectives under the {\it residuum} or R-conventions: the R-implication is defined as the residuum of the IMVL adopted t-norm via $ \nu(a \Longrightarrow_{_R} b) \equiv {\rm max}\{ z | \top(z, \nu(a)) \leq \nu(b) \}$ and the R-negation is defined from $\lnot_{_R} a \equiv a \Longrightarrow_{_R} 0$, where the zero-valuation on the right-hand side of the implication refers to a proposition false in all models. Finally, the R-disjunction is defined with respect to the corresponding t-conorm  $\nu(a \vee_{_R} b) \equiv \bot_{_R}(\nu(a),\nu(b)) \equiv 1-\top(\nu(\lnot_{_R}a),\nu(\lnot_{_R}b))$. Note that R-implications and R-disjunctions may be used with negations other than the R-negation \cite{Esteva}. 

There is a certain degree of freedom in how one goes about choosing a given t-norm. The rows 3-5 of the table shown in Fig. \ref{fig3} give the explicit expressions for all the connectives for three common t-norms: Godel, product and Lukasiewicz.

Interestingly, granting a truth value in the interval $[0,1]$ to logical propositions has led to further developments in set theory, pioneered by Zadeh \cite{Zadeh65}. The idea consists first in adopting a representation of traditional sets as follows. Consider a subset $A$ of a larger measurable set $\Omega$. The set $A$ can be characterised by its characteristic function $\chi_{A}$ acting on $\Omega$ such that for any $x \in \Omega$, $\chi_A(x)=1$ if $x \in A$ and $\chi_A(x)=0$ otherwise. The set $A$ can be represented as a realisation $\chi_A \in \{0,1\}^{\Omega}$ which assigns a value $0$ or $1$ to all elements $\Omega$. Zadeh then introduced the notion of {\it fuzzy sets} for which a given element in $\Omega$ can only partially belong to the set $A$ with a membership function $\mu_A$ returning a value between $0$ and $1$. In analogy to the previously introduced notation, fuzzy sets are then representable as a realisation $\mu_A \in [0,1]^{\Omega}$. On this view, traditional sets are then called {\it crisp sets} since the membership function can only return either $0$ or $1$. Typical operations between fuzzy sets such as complement, intersection, union and containment are then defined from their IMVL counterparts, {\it i.e.} t-norms, t-conorms and negation. In this way, fuzzy set theory enables vague linguistic variables such as short, tall, large and small, to be represented as fuzzy sets, so that inferences may be made about the relations between them. In its narrowest definition \cite{Dubois2007, Gottwald01}, a {\it fuzzy logic} is a logic that applies to fuzzy statements expressed as fuzzy sets, such that fuzzy inferences can be made. As such, fuzzy logic can be construed as an extension of IMVL with some additional structures that enable the handling of fuzzy notions. In what follows we will use {\it t-norm based fuzzy logics}, which are grounded in t-norm based IMVL and supplemented by some fuzzy rules. In Section \ref{sec3}, we shall rely on a specific fuzzy inference rule while in Section \ref{sec4} we shall briefly discuss the advantage that fuzzy variables may bring to the problem we are attempting to frame.

\section{Application of fuzzy logics to hypothesis rejection}
\label{sec3}
In the previous section we have introduced new tools with which it is possible to reason about partial truths. Within the generic MT argument for hypothesis rejection presented in Eqs. \eqref{eq2}-\eqref{eq4}, we can then try to assign a truth value to the premises ${\rm P1}$ and ${\rm P2}$.
Before doing so however, it is worth illustrating how probabilities and truth values differ from each other as a matter of principle. Probabilities, in their common acception, refer to the chances that propositions are exactly true. If we say that  "the probability for Tom to be tall is 0.3", then it usually means that we don't know Tom's height but that he belongs to an equivalence class of people among which 30\% are tall. If, on the other hand, one states that "the truth value for Tom to be tall is 0.3" then it means that Tom, as an individual, is only marginally tall. 
Notwithstanding this difference between truth valuations and probabilities, it is still the case that the former can be informed by the latter depending on the situation. In fact, if we interpret a truth value as a degree of confidence in a given proposition, then it can surely be influenced by probabilistic considerations if the said proposition has a statistical connotation. It is our understanding that ${\rm P1}$ carries a statistical meaning by asserting that, given $H$, the probability with which evidence $E$ is observed is larger than some value $\alpha$. This is because, to some extent, ${\rm P1}$ can be interpreted as a proposition about the future and, following Carl Friedrich von Weizsacker, the future is deeply connected to probabilities \cite{Drieschner14}. Of course, if $\alpha =0 $ then ${\rm P1}$ is a tautology because the consequent $P(E|H) \geq 0$ is always true. On that basis, it is tempting to suggest that the truth value of ${\rm P1}$ is bound from below by the probability of the consequent to occur. Assuming the observed evidence is sampled from the conditional probability measure, this probability is $1-\alpha$ \footnote{\label{f1}This can be seen in the case of observed evidence taking the form $E \equiv S \geq s$. In that case $P(S \geq s|H)$ is connected to a cumulative distribution and $P(S \geq s|H) \geq \alpha$ is a condition on it. Given that a cumulative distribution is monotonous, it can be translated onto a condition for $s$ as $s \geq P^{-1}(\alpha | H)$. Assuming a probability measure $\mu_{H}(s)$ over $s$ we get that the measure of the set of points $s$ for which the condition is satisfied is $\int_{P^{-1}(\alpha | H)}^{+\infty} d\mu_{H}(s) =\int_0^{+\infty} d\mu_{H}(s) - \int^{P^{-1}(\alpha | H)}_{0} d\mu_{H}(s) = 1-P(P^{-1}(\alpha|H)|H)= 1-\alpha $ }. Therefore, we may posit that the truth value for ${\rm P1}$ satisfies $\nu({\rm P1}) \geq 1-\alpha$. Note that upon conservatively choosing $\nu({\rm P1}) = 1-\alpha$ for the valuation of ${\rm P1}$, we do obtain truth values different from unity whenever $\alpha > 0$: the difference is now that there is no direct collapse into complete falsity. This intuition can be made more formal when trying to assign a probability to a conditional statement. Nguyen et al. \cite{Nguyen02} have proposed an interpolating expression between logical truth valuation function (which they call logical probability) and statistical probability for which the minimum valuation for a conditional $\nu(A \Longleftarrow B) = P(A|B)$, which is equal to $P(E > P^{-1}(\alpha|H)|H) = 1-\alpha$ in our case (see footnote \ref{f1}).
Contrary to ${\rm P1}$ the valuation for ${\rm P2}$ is fairly straightforward since it concerns actually observed evidence. So, given an a priori probability $P(E|H)$ and an observed evidence $E$ (which we consider infinitely accurate for now) such that $P(E|H) < \alpha$, there is no particular doubt about the truthiness of the premise ${\rm P2}$. As a result we assign a truth value $\nu({\rm P2}) = 1$ to ${\rm P2}$. Note again the striking difference with the statistical interpretation: if ${\rm P2}$ were to be interpreted statistically, there would usually be some very low probability to observe some evidence with conditional probability lower than $\alpha$. In fuzzy logic, however, we instead assign a truth value to an actual observation $E$ and a mathematical outcome $P(E|H) < \alpha$ following this observation; hence the truth value of 1. Finally, it is important to stress that $1-\alpha$ has been assigned to ${\rm P1}$ and {\it not} to the consequent $P(E|H) \geq \alpha$. Indeed, for the overall argument to be consistent, the truth value of $P(E|H) \geq \alpha$ has to be connected to our choice that $\nu({\rm P2}) = 1$.

To finish discussing how MT can be used within fuzzy logic, we need to appreciate that because the premises can in principle have any value between $0$ and $1$, so does the conclusion. However, since MT is a rule and not a connective, we cannot assign a value to this conclusion from the table shown in Fig. \ref{fig3} alone. Instead, we need to posit an extension of MT which preserves some of the desirable properties of such an inference rule \cite{Smets89}. In what follows we will use Zadeh's {\it compositional extension} which, for MT, prescribes that the truth value of the conclusion $\lnot H$ is

\begin{equation}
 \nu(\lnot H) \equiv \top \left( \nu( \lnot [P(E|H) \geq \alpha] \Longrightarrow \lnot H), \nu(P(E|H) < \alpha) \right).  \label{eq5}
 \end{equation}
To evaluate Eq. \eqref{eq5} we need to distinguish two possible cases: whether one is dealing with S-Implications or R-Implications. The reason lies in the fact that the first argument of the t-norm in Eq. \eqref{eq5} is the valuation of the sentence  $\lnot [P(E|H) \geq \alpha] \Longrightarrow \lnot H$ which is the contrapositive of ${\rm P1}$. Depending on the kind of implication being used, extracting $\nu( \lnot [P(E|H) \geq \alpha] \Longrightarrow \lnot H)$ from $\nu({\rm P1})$ is not trivial.

\vspace{3mm}

\textbf{Modus Tollens under S-Implications in fuzzy logic}

\vspace{2mm}
Under an S-Implication with a provided (canonical) negation $\nu(\lnot_S a)\equiv 1-\nu(a)$ we get that for any proposition $a$ and $b$ we have:
\begin{eqnarray}
 \nu( \lnot_S b \Longrightarrow_S \lnot_S a ) &\equiv& \bot(\nu(\lnot_S \lnot_S b), \nu(\lnot_S a)) = \bot(\nu( b), \nu(\lnot_S a)) \label{eq6} \\
 &=& \bot(\nu(b), 1-\nu(a)) = \bot(\nu(b),1)-\bot(\nu(b),\nu(a))= \bot(1,\nu(b))-\bot(\nu(a),\nu(b)) \label{eq7} \\
 &=& \bot(\nu(\lnot_S a), \nu(b)) \equiv  \nu( a \Longrightarrow_S b ) \label{eq8}
\end{eqnarray}where we used the fact that $\lnot_S \lnot_S a \equiv a$ in Eq. \eqref{eq6} and both the distributivity and permutation symmetry of the t-conorm were used in Eq. \eqref{eq7}.\newline Eq. \eqref{eq8} is called {\it contrapositive symmetry} and shows that it holds for any S-Implication.

\vspace{2mm}

From this contrapositive symmetry we get that Eq.\eqref{eq5} reads

\begin{equation}
 \nu(\lnot_S H) = \top(\nu({\rm P1}),\nu({\rm P2}) ) = \top(1-\alpha, 1) = 1-\alpha \label{eq9}
\end{equation}for all t-norms. For the last equality of Eq.\eqref{eq9} we used axiom 4 of t-norms which identifies 1 as the neutral element.

\vspace{2mm}

At this stage, it is worth checking whether the MT inference derived above is internally consistent {\it i.e} whether or not the conclusion contradicts some of the premises. One can see quite quickly that it is possible to assign a non-zero truth value to an S-Implication even if the consequent is exactly false (which is the case here because $\nu({\rm P2})=1$). In fact for the three models of t-norms we have chosen ({\it i.e.} Godel, product and Lukasiewicz) the outcome of the S-Implication in ${\rm P1}$ for a false consequent is $1-\nu( H)$. Since we have assigned $\nu({\rm P1})=1-\alpha$, it follows that $\nu(H) = \alpha$. Under the negation rule we have used we get $\nu(\lnot_S H) = 1-\nu(H) = 1-\alpha$; which is consistent with the final result of Eq. \eqref{eq9}.

\vspace{3mm}

\textbf{Modus Tollens under R-Implications and R-negations in fuzzy logic}

\vspace{2mm}

For R-Implications we are going to see that the consistency of the proposed argument depends on the chosen t-norm.

\begin{itemize}
\item Lukasiewicz t-norm: as can be seen from the table shown in Fig. \ref{fig3}, the R-negation for this t-norm is the canonical negation $\nu(\lnot_R a)=1-\nu(a)$. Applying this to the truth value of the contrapositive to an implication of the form $ a \Longrightarrow_R b$ we get
\begin{eqnarray}
\nu( \lnot_R b \Longrightarrow_R \lnot_R a) &\equiv& {\rm min}(1,1-\nu(\lnot_R b)+\nu(\lnot_R a)) \label{eq10} \\
&=& {\rm min}(1,1-(1-\nu(b))+(1-\nu(a))) = {\rm min}(1,1-\nu(a)+\nu(b)) \label{eq11} \\
&\equiv& \nu(a \Longrightarrow_R b) \label{eq12}
\end{eqnarray}which shows contrapositive symmetry. As a result, the MT inference in Eq. \eqref{eq9} assigning a truth value $\nu(\lnot_R H) = 1-\alpha$ is valid for this R-Implication.

\vspace{2mm}

\item Godel and product t-norms: from the table shown in Fig. \ref{fig3} we see that the negation for Godel and product t-norms still entail that $\nu(P(E|H) \geq \alpha) = 0$ if we have $\nu({\rm P2})=1$. What remains to be assessed is whether such a result leads to a compatibility between our evaluation that $\nu({\rm P1})=1-\alpha$ and the R-Implication rule for Godel's and products t-norms. In fact that is not the case. Both R-Implications will in fact return $\nu({\rm P1}) = 0$ if one assumes $\nu(H) > 0$ or $\nu({\rm P1}) = 1$ if $\nu(H) = 0$. In either case, this is not compatible with the probability-informed truth value estimate of $1-\alpha$ that we have suggested for premise ${\rm P1}$. 
\end{itemize}

In the end, while the Lukasiewicz R-implication gives an identical result to S-implications, it is not possible to conduct the MT inference argument from Eqs. \eqref{eq5} with the Godel and product t-norms without running into consistency issues. In fact the induced R-negation for such t-norms drastically restricts the acceptable valuation models for ${\rm P1}$ and ${\rm P2}$ ({\it e.g} it is not possible to have $0< \nu({\rm P2}) < 1$). One may argue that such restrictions defeat the whole point of using fuzzy logics in the first place, but in fact they simply model different facets of the corresponding classical connectives. For example, a classical conditional is automatically false if the consequent is false and it is conceivable to represent a negation as an operation making exactly false all partial truths. Nonetheless, it is worth exploring how the R-Implications behave when used with the S-negation. 

\vspace{3mm}

\textbf{Modus Tollens under R-Implications and S-negation in fuzzy logic}

\vspace{2mm}

We now look at fuzzy logics using R-implications alongside the S-negation. Our result on the Lukasievicz R-implication is unaffected by such a change since its corresponding R-negation was already the S-negation. We will therefore focus on the problematic ones: the Godel and product R-implications. First, we need to recognise a problem that the residua of each of these t-norms has: if the consequent has truth value zero, it is not possible to obtain a non-zero truth value for their corresponding R-implications and therefore impossible to have $\nu({\rm P1}) > 0$. To obtain a non-zero truth value for the consequent in ${\rm P1}$, we need to relax the absolute truth value of ${\rm P2}$. Assuming there is some precision --- usually very small compared to the signal itself --- in any measurement, we shall posit more generally that $\nu({\rm P2})=1-p$ where $p$ represents the error due to the limited precision. From the S-negation it follows that $\nu(P(E|H)\geq \alpha)=p$    \footnote{Note that with the Godel and Product R-negation, the result would have been strictly zero instead.}.   We can now try to see how this might affect our previous conclusions on the use of the Godel and product R-implications in MT.

\vspace{2mm}

\begin{itemize}
\item Godel t-norm: from the table shown in Fig. \ref{fig3} we see that if the consequent of the R-implication has non-zero truth value $p$, the truth value of the implication itself can only be either $1$ if $\nu(H) \leq p$ or $p$ otherwise. The former possibility cannot model a situation where $\nu({\rm P1}) = 1-\alpha$ (for $0< \alpha \leq 1 $) so the only compatible model is that the R-implication (and therefore $\nu({\rm P1})$) has truth value $p$. However, this requirement is inconsistent with the fact that $p$ is {\it a priori} independent of $\alpha$ since they correspond to two very different statements. For the Godel t-norm, using an S-negation and relaxing the certainty on ${\rm P2}$ is therefore not sufficient.
\vspace{2mm}
\item Product t-norm: from the table shown in Fig. \ref{fig3} we see that if the consequent has valuation $p$ the possible valuations of the implication itself are either $1$ if $\nu(H) \leq p$ or $p/\nu(H)$ otherwise. Of these two possible cases, only the second one is compatible with admitting a valuation $1-\alpha$ for ${\rm P1}$. Requiring consistency compels us to get 
\begin{equation}
\nu(H) = {\rm min}\left(1,\frac{p}{1-\alpha}\right) \label{eq13}
\end{equation}
where the min function ensures that $\nu(H) \leq 1$. Having used consistency to find a truth value to $H$ we can check whether or not it is consistent with the valuation proposed in Eq. \eqref{eq5}. To do so, we need first to evaluate the truth value of the contrapositive of a generic R-implication:
\begin{equation}
\nu( \lnot_S b \Longrightarrow_R \lnot_S a) \equiv {\rm min}\left(1, \frac{1-\nu(a)}{\nu(\lnot_S b)} \right) = {\rm min}\left(1, \frac{1-\frac{1-\nu(\lnot_S b)}{\nu( a \Longrightarrow_R b)}}{\nu(\lnot_S b)} \right) \label{eq14}
\end{equation}where the last equality is obtained from the product R-implication under the condition that $\nu(a \Longrightarrow_{_R} b) < 1$. We can then substitute into the compositional extension rule for MT:
\begin{eqnarray}
\nu(\lnot_S a) &=& \top(\nu( \lnot_S b \Longrightarrow \lnot_S a) , \nu(\lnot_S b)) = \nu(\lnot_S b) \times {\rm min}\left(1, \frac{1-\frac{1-\nu(\lnot_S b)}{\nu( a \Longrightarrow_R b)}}{\nu(\lnot_S b)} \right) \nonumber \\
&=& {\rm min} \left( \nu(\lnot_S b), 1-\frac{1-\nu(\lnot_S b)}{\nu(a \Longrightarrow_R b)} \right)  \nonumber \\
&=& 1-\frac{1-\nu(\lnot_S b)}{\nu(a \Longrightarrow_R b)} \label{eq15}
\end{eqnarray}where the last equality follows from the requirement that $\nu(a \Longrightarrow_R b) < 1$ and the S-negation. Substituting $\nu(\lnot_S b)$ for $1-p$ and $\nu(a \Longrightarrow_R b)$ for $1-\alpha$ in Eq. \eqref{eq15} gives us the same result as Eq. \eqref{eq13} under S-negation.
\end{itemize}
Upon using a logic mixing R-implications and S-negation we were then able to lift some issues reported for the product t-norm with R-implications and R-negations. It is worth noting that these changes were not enough to make the MT reasoning with valuation from Eq. \eqref{eq5} compatible with Godel's t-norm.

\section{Discussion and conclusion}
In this work we have reminded and illustrated the fact that SHT, as it is typically practiced today, is logically flawed. On the one hand Bayesian inference would require two unknown parameters in order to deliver any judgement while, on the other hand, the MT logical skeleton mixed with probabilistic vocabulary makes it an unsound argument within classical logic. To look for a logic on which to ground the intuition behind $P$ value based hypothesis rejection, we have explored t-norm based fuzzy logics under the S- and R- conventions for the three most commonly employed t-norms. We have found that: $(a)$ a fuzzy extension of MT with the S-implication and S-negation constitute a viable logical framework on which to ground frequentist SHT for all t-norms; $(b)$ when R-implications and R-negations are used, only the Lukasiewicz t-norm can be used for MT while the Godel and product t-norms yield results incompatible with the suggested model; and $(c)$ upon using the S-negation with R-implications and relaxing some assumptions of the model, it was possible to salvage the product t-norm as a possible framework for SHT. These findings suggest that fuzzy logic may be suitable to ground SHT and reflect deeper on the kind of assumptions (including logical axioms, t-norms etc...) necessary for such reasonings to be valid and sound. For example, one may wonder what is so special about fuzzy logic that it appears to bypass the need for the two supplementary parameters that a Bayesian inference would require (see Fig. \ref{fig1}). Does this mean that there is {\it free lunch}? Our current understanding of the fuzzy reasoning which have shown promise ({\it i.e.} S-implications and S-negation) suggests that this {\it free lunch} has two origins: the existence of a robust contrapositive symmetry (derived in Eq.\eqref{eq8}) on the one hand and the replacement of statistical probability (about ensembles) by the truth value judgement about an event which has actually occurred ({\it e.g.} for stating that $\nu({\rm P2})=1$) on another hand. Such properties do not exist in statistical reasonings and it is not necessarily desirable that they do; after all, although probabilities and truth values take values between 0 and 1, they are ultimately different things.

If one chooses $\alpha = 0.05$ and has no doubt about the fact that the $P$ value is lower than $0.05$, then Eq.\eqref{eq9} says that it can be (fuzzy) logically concluded that the hypothesis is false with truth value $0.95$. This supports current practice in medical research. It is essential to note that $0.95$ is not a probability and, more importantly, that this result strongly depends on the valuations we proposed for ${\rm P1}$ and ${\rm P2}$. While we do believe that we have provided a sensible rationale for these truth values, fuzzy logic alone doesn't impose a direction on what they should be. A more radical modelling approach for $\nu({\rm P1})$ could have been to remain as close as possible to the classical strictly false value for ${\rm P1}$ by allowing a deviation from zero which depends only slightly on the value of $\alpha$. For example, any valuation of the form $\nu_n({\rm P1}) \equiv 1-\alpha^n$ with $n > 0$ would be perfectly legitimate as far as limiting cases are concerned but would give rise to a drastically different truth value for the conclusion. In that regard, far from being final, we believe the present work opens up research avenues in the application of (fuzzy) logic to the justification and critical appraisal of scientific practice: one, as mentioned before, in the modelling of valuation functions for the premises in the MT inference rule and the other related to the fundamental problem that it is not clear what constitutes a "small enough" value for $\alpha$ to warrant actual decision making. If one can reject a hypothesis with truth value $0.95$, to what extent would it be legitimate to reject it if the truth value was instead $0.9$? For instance, particle physics works with $P$ values of the order of $3 \times 10^{-7}$ (one-tailed test) with the 5 sigma convention, which has attracted attention in relation to the announcement of the Higgs boson detection in 2012. These discrepancies show that the whole $P$ value reasoning is based on a somewhat vague concept of "small enough": "if the $P$ value is "small enough" then we can reject the hypothesis". Incidentally, fuzzy logic offers tools which we haven't had the opportunity to use in the present paper but which would nonetheless enable one to deal with such vagueness. These are the so-called {\it linguistic variables} \cite{Zadeh08} which replace the strict propositions to be evaluated ({\it e.g.} "$P$ value less than $0.05$") by vague sentences modelled as fuzzy sets ({\it e.g.} "$P$ value is small"). 

In the end, we have showed that, as a proof of principle, it was possible to ground SHT in fuzzy logic and model current practices of frequentist statistics. Far from being the final word on the matter, we hope that this article will spur further works along these lines to delineate the kind of reasonings being used in scientific discourses. 
\label{sec4}
\bibliography{biblio}

\begin{thebibliography}{10}

\bibitem{Chalmers1999}
A.F. Chalmers.
\newblock {\em What Is This Thing Called Science?}
\newblock Open University Press, 3rd edition, 19999.

\bibitem{Consonni18}
G.~Consonni, D.~Fouskakis, B.~Liseo, and I.~Ntzoufras.
\newblock Prior distributions for objective bayesian analysis.
\newblock {\em Bayesian Analysis}, 13:227--279, 2018.

\bibitem{Drieschner14}
M.~Drieschner, editor.
\newblock {\em Carl Friedrich von Weizsacker: Major texts in Physics}.
\newblock Springer, 2014.

\bibitem{Dubois2007}
Didier Dubois, Francesc Esteva, Lluís Godo, and Henri Prade.
\newblock Fuzzy-set based logics ? an history-oriented presentation of their
  main developments.
\newblock In D.~M. Gabbay and J.~Woods, editors, {\em Handbook of the Hitsory
  of Logic 8}. Elsevier, 2007.

\bibitem{Duhem}
P.~Duhem.
\newblock {\em The Aim and Structure of Physical Theory}.
\newblock Princeton University Press, 1954.

\bibitem{Esteva}
Francesc Esteva, Lluis Godo, Petr Hajek, and Mirko Navara.
\newblock Residuated fuzzy logic with an involutive negation.
\newblock {\em Arch. Math. Log.}, 39:103--124, 2000.

\bibitem{Fisher55}
R.~Fisher.
\newblock Statistical methods and scientific induction.
\newblock {\em Journal of the Royal Statistical Society. Series B
  (Methodological)}, 17(1):69--78, 1955.

\bibitem{Gigerenzer04}
G.~Gigerenzer.
\newblock Mindless statistics.
\newblock {\em The Journal of Socio-Economics}, 33(5):597--606, 2004.

\bibitem{Goodman99}
S.~N. Goodman.
\newblock Towards evidence-based medical statistics. 1: The p value fallacy.
\newblock {\em Annals of Internal Medicine}, 130(12):995--1004, 1999.

\bibitem{Gottwald01}
S.~Gottwald, editor.
\newblock {\em A treatise on many-valued logics}.
\newblock Research Studies Press, 2001.

\bibitem{Greco11}
D.~Greco.
\newblock Significance testing in theory and practice.
\newblock {\em Brit. J. Phil. Sci.}, 62:607--637, 2011.

\bibitem{Lehmann1993}
E.~L. Lehmann.
\newblock The fisher, neyman-pearson theories of testing hypotheses: One theory
  of two?
\newblock {\em American Statistical Association}, 88:201--208, 1993.

\bibitem{Lu20}
Chenguang Lu.
\newblock Channels? confirmation and predictions? confirmation: From the
  medical test to the raven paradox.
\newblock {\em Entropy}, 22(4), 2020.

\bibitem{Smets89}
P.~Magrez and P.~Smets.
\newblock Fuzzy modus ponens: A new model suitable in knowledge based systems.
\newblock {\em International Journal of Intelligent Systems}, 4:181--200, 1989.

\bibitem{Neyman1928}
J.~Neyman and E.S. Pearson.
\newblock On the use and interpretation of certain test criteria for purposes
  of statistical inference: Part i.
\newblock {\em Biometrika}, 20A:175--240, 1928.

\bibitem{Neyman1933}
J.~Neyman and E.S. Pearson.
\newblock Ix. on the problem of the most efficient tests of statistical
  hypotheses.
\newblock {\em Phi. Trans. R. Soc. Lond. A.}, 231:289--337, 1933.

\bibitem{Nguyen02}
H.~T. {Nguyen}, M.~{Mukaidono}, and V.~{Kreinovich}.
\newblock Probability of implication, logical version of bayes theorem, and
  fuzzy logic operations.
\newblock In {\em 2002 IEEE World Congress on Computational Intelligence. 2002
  IEEE International Conference on Fuzzy Systems. FUZZ-IEEE'02. Proceedings
  (Cat. No.02CH37291)}, volume~1, pages 530--535 vol.1, 2002.

\bibitem{Nickerson00}
R.S. Nickerson.
\newblock Null hypothesis significance testing: a review of an old and
  continuing controversy.
\newblock {\em Psychol. Methods}, 5:241--301, 2000.

\bibitem{Perezgonzalez15}
Jose~D. Perezgonzalez.
\newblock Fisher, neyman-pearson or nhst? a tutorial for teaching data testing.
\newblock {\em Frontiers in Psychology}, 6:223, 2015.

\bibitem{Popper}
K.~Popper, editor.
\newblock {\em The Logic of Scientific Discovery}.
\newblock Routledge, 2005.

\bibitem{Priest10}
G.~Priest.
\newblock The logic of catuskoti.
\newblock {\em Comparative Philosophy}, 1(2):24--54, 2010.

\bibitem{Smets87}
P.~Smets and P.~Magrez.
\newblock Implication in fuzzy logic.
\newblock {\em International Journal of Approximate Reasoning}, 1:327--347,
  1987.

\bibitem{Sober1999}
E.~Sober.
\newblock Testability.
\newblock {\em Proceedings and Addresses of the American Philosophical
  Association}, 73:47--76, 1999.

\bibitem{Stern14}
J.~M. Stern and C.~A. De~Braganca~Pereira.
\newblock Bayesian epistemic values: focus on surprise, measure probability!
\newblock {\em Logic Journal of the IGPL}, 22:236--254, 2014.

\bibitem{Stern17}
J.~M. Stern, R.~Izbicki, and R.~B. Esteves, L. G.~Stern.
\newblock Logically-consistent hypothesis testing and the hexagon of
  oppositions.
\newblock {\em Logic Journal of the IGPL}, 25:241--757, 2017.

\bibitem{Tendeiro19}
J.~N. Tendeiro and H.~A.~L. Kiers.
\newblock A review of issues about null hypothesis bayesian testing.
\newblock {\em Psychological Methods}, 24:274--795, 2019.

\bibitem{Zadeh65}
L.~A. Zadeh.
\newblock Fuzzy sets.
\newblock {\em Information and Control}, 8:338--353, 1965.

\bibitem{Zadeh08}
L.~A. Zadeh.
\newblock Is there a need for fuzzy logic?
\newblock {\em Information Sciences}, 178:2751--2779, 2008.

\end{thebibliography}
\bibliographystyle{plain}

\end{document}